\documentclass[letterpaper, 10 pt, conference]{ieeeconf}  % Comment this line out if you need a4paper

\IEEEoverridecommandlockouts                              % This command is only needed if 
\usepackage{graphics} % for pdf, bitmapped graphics files
\usepackage{subfig}
\usepackage{epsfig} % for postscript graphics files
\usepackage{mathptmx} % assumes new font selection scheme installed
\usepackage{times} % assumes new font selection scheme installed
\usepackage{amsmath} % assumes amsmath package installed
\usepackage{amssymb}  % assumes amsmath package installed
\usepackage{color}
\usepackage{soul}
\usepackage[normalem]{ulem}
\usepackage{lipsum}
\usepackage[disable]{todonotes}

\newcommand{\comment}[1] {} %comment showed
\usepackage{amsmath} 
\usepackage{adjustbox}
\usepackage{multirow}
\usepackage{comment}
\usepackage[colorlinks,linkcolor=black]{hyperref}

\usepackage{algorithm}
\usepackage[noend]{algpseudocode}

\usepackage{array}

\usepackage{parselines}

\title{\LARGE \bf Behavior Planning at Urban Intersections through Hierarchical Reinforcement Learning$^*$
}

\author{Zhiqian Qiao$^{1}$, Jeff Schneider$^{2}$ and John M. Dolan$^{2}$ % <-this % stops a space
\thanks{*This work is supported by Argo AI} % <-this % stops a space
\thanks{$^{1}$Zhiqian Qiao is a Ph.D. student of Electrical and Computer Engineering, Carnegie Mellon University, 5000 Forbes Ave, Pittsburgh, USA
        {\tt\small zhiqianq@andrew.cmu.edu}}%
\thanks{$^{2}$ The Robotics Institute, Carnegie Mellon University}
}

\begin{document}

\maketitle
\thispagestyle{empty}
\pagestyle{empty}

%%%%%%%%%%%%%%%%%%%%%%%%%%%%%%%%%%%%%%%%%%%%%%%%%%%%%%%%%%%%%%%%%%%%%%%%%%%%%%%%
\begin{abstract}

  For autonomous vehicles, effective behavior planning is crucial to ensure safety of the ego car. In many urban scenarios, it is hard to create sufficiently general heuristic rules, especially for challenging scenarios that some new human drivers find difficult. In this work, we propose a behavior planning structure based on reinforcement learning (RL) which is capable of performing autonomous vehicle behavior planning with a hierarchical structure in simulated urban environments. Application of the hierarchical structure allows the various layers of the behavior planning system to be satisfied. Our algorithms can perform better than heuristic-rule-based methods for elective decisions such as when to turn left between vehicles approaching from the opposite direction or possible lane-change when approaching an intersection due to lane blockage or delay in front of the ego car. Such behavior is hard to evaluate as correct or incorrect, but for some aggressive expert human drivers handle such scenarios effectively and quickly. On the other hand, compared to traditional RL methods, our algorithm is more sample-efficient, due to the use of a hybrid reward mechanism and heuristic exploration during the training process. The results also show that the proposed method converges to an optimal policy faster than traditional RL methods.

\end{abstract}

%%%%%%%%%%%%%%%%%%%%%%%%%%%%%%%%%%%%%%%%%%%%%%%%%%%%%%%%%%%%%%%%%%%%%%%%%%%%%%%%
\section{INTRODUCTION}

An autonomous vehicle (AV) consists of three main systems: 1) Perception, 2) Planning and 3) Control. By analogy with a human driver, perception can be compared to the human's eyes, planning to the human's brain and consciousness, and control to the human's arms and legs. Generally, the whole driving system needs the sensors or human eyes to detect the environment and localize the vehicle. Then according to the map information and the prediction of the events happening, the route planner gives out the route from starting point to destination. The behavior planner then makes behavior-level decisions for driving like merging onto the exit lane, turning left, etc. Then the autonomous vehicle needs to generate the trajectory for the controller to follow, while human drivers can control the car by limbs through feelings and experience directly. The human brain and consciousness are mysterious in many ways and contain various phenomena and characteristics that are difficult to explain. Similarly, the planning part of driving is difficult, especially for new human drivers, due to lack of experience. 

\begin{figure}[!t]
  \centering
  \includegraphics[width=\columnwidth]{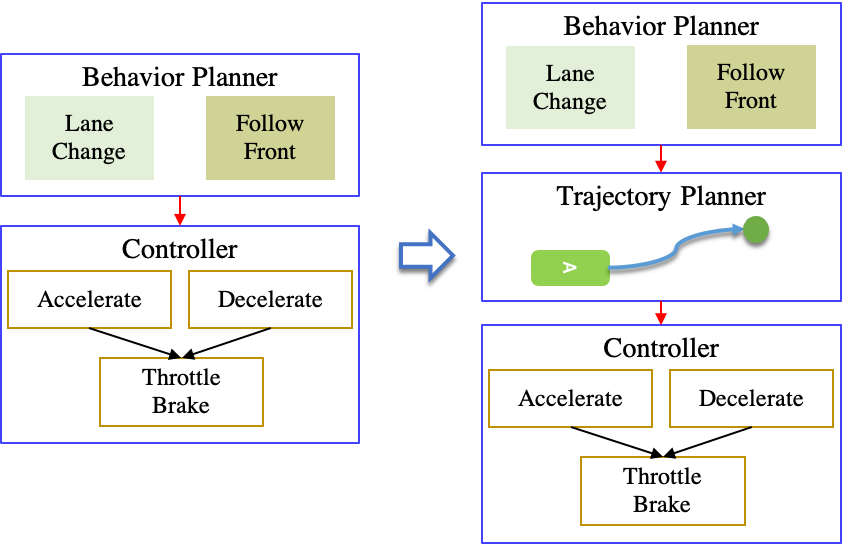}
  \caption{Hierarchical structure of planning system with hierarchical reinforcement learning}
  \label{fig_hrl}
  \vspace{-0.5cm}
\end{figure}

Understanding how human drivers make decisions while driving can provide heuristics for autonomous vehicles. The behavioral approach towards decision-making emphasizes that the decision makers are prone to varying degrees of rationality while making decisions, which makes it hard to ensure complete rationality at all times. In order to mimic this kind of human behavior, we propose a hierarchical reinforcement learning structure which can generate both higher-level behavior decisions and a lower-level controller. However, applying the controller by directly connecting to the behavior decision with reinforcement learning causes jerky and unsafe maneuvers for some complicated urban intersection scenarios, especially when the steering angle, throttle and brake are controlled together. 

A traditional planner structure for a self-driving car consists of a route planner, behavior planner and trajectory planner. In this work, the hierarchical structure for the behavior planning system with RL is shown in Figure \ref{fig_hrl}. The two-layer setup includes a high-level behavior planner and a low-level controller. For a traditional planning system, the trajectory planner can increase the car's control stability. As a result, in this work we also propose a three-layer setup which adds an intermediate layer in order to output the target trajectory points. The system would first output the high-level behavior decision. Then, depending on the decision, the trajectory layer generates a corresponding target waypoint that the ego car intends to follow. Finally, the action layer outputs the throttle, brake and steering angle. Adding a trajectory layer to the hierarchical reinforcement learning structure has the following advantages: 1. it avoids the exploration of some obviously unsafe behaviors during the training process; 2. the low-level controller is more smooth in following the target waypoints received from the trajectory layer. This paper's contributions are:
\begin{enumerate}
    \item A hierarchical RL structure with different layers for the planning system of a self-driving car.
    \item Application of a hierarchical behavior planning structure at urban intersections.
    \item Demonstration of improved performance of our algorithms compared to traditional heuristic rule-based methods and vanilla RL algorithms, using various metrics.
\end{enumerate}

\section{Related Work}

This section summarizes previous work related to this paper under the following headings: 1) papers that propose self-driving behavior planning algorithms; 2) papers that address reinforcement learning (RL) technologies.

\subsection{Behavior Planning of Autonomous Vehicles}

Previous work applied heuristic-based rules-enumeration, imitation learning and  reinforcement learning algorithms to the behavior planning of autonomous vehicles based on different scenarios.

Time-to-collision (TTC) \cite{ttc} is a typical heuristic-based algorithm that is used to be compared with as a baseline for various learning-based algorithms. \cite{slot} proposed a slot-based approach to check if it is safe to merge into lanes or across an intersection with moving traffic. This method is based on the information of slots available for merging, which includes the size of the slot in the target lane, and the distance between the ego-vehicle and front vehicle. However, the proposed heuristic-based methods rely heavily on the parameter tuning and each set of parameters is restricted to the corresponding scenarios and environments. The use of heuristics alone makes it hard to make the algorithm sufficiently general when designing a high-performance autonomous vehicle behavior planning system. Especially for complex urban scenarios, it is laborious and time-consuming to develop a set of rules with or without advanced technology which can cover all possible cases.

Imitation learning is an alternative method for a self-driving behavior planner that requires a large amount of data collected from human expert drivers. The fundamental imitation learning algorithms can be categorized into behavior cloning \cite{ross2011reduction}\cite{zhang2016query}\cite{sun2018fast}, direct policy learning and inverse reinforcement learning (IRL) \cite{irl}. \cite{chen2015deepdriving} and \cite{bojarski2016end} are examples of using behavior cloning in order to mimic human driver data collected from real-world cars or high-fidelity simulations. This kind of supervised learning algorithm can work well if the state-action pairs during testing are similar to that during training and the assumption is met that the state-action pair are independent and identically distributed. However, for most driving scenarios, the driving process is a Markov Decision Process in which the current state relies on the previous state, so that the error for generating an action will continuously increase for an action generated by behavior cloning only. Direct policy learning. \cite{dorsaplanning} modeled the interaction between autonomous vehicles and human drivers by the method of IRL in a simulated environment. However, IRL needs two processes of reward learning and reinforcement learning, which leads to difficult convergence during the training process. 

If the reward function is not difficult to get from experts or it can be easily gotten from the heuristic rules described before, reinforcement learning \cite{rl}\cite{dqn}\cite{ddpg}\cite{doubledqn} is easier to apply to the self-driving problem. RL is capable to transfer multiple rules into a mapping function or neural network. \cite{pomdpdecision} formulated the decision-making problem for AV under uncertain environments as a POMDP and trained out a Bayesian Network to deal with a T-shape intersection merging problem. \cite{navigateintersection} used Deep Recurrent Q-network (DRQN) with states from a bird's-eye view of the intersection to learn a policy for traversing the intersection. \cite{occludednavigating} proposed an efficient strategy to navigate through intersections with occlusion by using the DRL method. These works focused on designing variants of the state-space and add-on network modules in order to allow the agent to handle different scenarios. However, RL has difficulties to validate the intermediate process. For the self-driving car scenarios, the planning system needs Facilitate validation and explainable behavior to ensure the safeness. In this work, we proposed to build a hierarchical learning-based structure which allow the validation of different local policies as a sub-function with fully capabilities within the hierarchical system instead of presenting a monolithic neural-network black-box policy.

\subsection{Reinforcement Learning}

Algorithms with extended functions based on RL and Hierarchical RL \cite{maxq}\cite{rmax}\cite{hrl} have been proposed. For the hierarchical structure, \cite{hdrl} proposed the idea of a meta controller, which is used to define a policy governing when the lower-level action policy is initialized and terminated. \cite{parl} used the idea of the hierarchical model and transferred it into parameterized action representations. Most previous work designs a single hierarchical structure which can be used to solve the entire problem. However, in most real-world cases, a complicated task such as behavior planning of autonomous vehicle can be compromised with several sub-tasks.

Some previous work are using human expert data to deal with the exploration problem during the RL training process. For example, \cite{hester2017deep} proposed the Deep Q-learning from Demonstrations (DQfD) which used the demonstration data collected from human to accelerate the learning process. \cite{nair2018overcoming} use demonstrations to improve the exploration process and successfully learn to perform long-horizon, multi-step robotics tasks with continuous control by using Deep Deterministic Policy Gradients and Hindsight Experience Replay. In our work, instead of using demonstration data directly, we include the heuristic-based rules-enumeration policy to during the exploration process for multiple hierarchical planning layers in order to accelerate the training process massively. We proposed to build the HRL-structure according to the heuristic method so that the system can adjust the exploration rate according to the training results real time and meanwhile can more easily figure out the local optimal policy based on the environment.

\section{Methodology}
In this section we present our proposed model, which is a HRL-based planning system with a hybrid reward mechanism and an adjusted heuristic exploration training schema. Meanwhile, we introduce the behavior and trajectory planning system. We will refer to this model as HybridHRL throughout the paper.

\subsection{Preliminaries}

Firstly, we introduce the fundamentals of the Hybrid hierarchical reinforcement learning (HybridHRL) method, which is based on DQN \cite{dqn} and DDQN \cite{doubledqn}. These two proposed methods have been widely applied in various reinforcement learning problems. For Q-learning, an action-value function $Q_{\pi}(s, a )$ is learned to get the optimal policy $\pi$ which can maximize the action-value function $Q^*(s, a)$:
\begin{equation}
    Q^*(s, a)  = \max_{\mathbf{\theta}} Q(s, a | \mathbf{\theta})  = r + \gamma \max_{\mathbf{\theta}} Q(s', a' | \theta)
\end{equation}
$s$ and $a$ are current state and action, respectively. When updating the network, for DQN, the loss function and updating methods can be written as:
\begin{equation}
    \label{equ_dqnL}
    \begin{split}
        Y_t^Q  & = R_{t+1} + \gamma \max_a Q(S_{t+1}, a | \mathbf{\theta_t})\\
        L(\mathbf{\theta}) & = \left(Y_t^Q -  Q(S_t, A_t | \mathbf{\theta}_t ) \right)^2\\
        \mathbf{\theta}_{t+1} & = \mathbf{\theta}_{t} + \alpha \frac{\partial L(\mathbf{\theta})}{\partial \mathbf{\theta}}\\
    \end{split}
\end{equation}

For the DDQN, the target action-value $Y^Q$ is revised according to another target Q-network $Q^{'}$ with parameter $\mathbf{\theta}'$:
\begin{equation}
    \label{equ_ddqn}
    \begin{split}
        Y_t^Q  & = R_{t+1} + \gamma Q(S_{t+1}, \arg\max_a Q'(S_{t+1}, a | \mathbf{\theta}_t) | \mathbf{\theta}_t')
    \end{split}
\end{equation}

For the original HRL model \cite{hdrl} with sequential sub-goals, a meta controller $Q^1$ generates the sub-goal $g$ for the following steps and a controller $Q^2$outputs the actions based on this sub-goal until the next sub-goal is generated by the meta controller. The objective functions are chosen separately: 
\begin{equation}
    \label{equ_hq}
    \begin{split}
        Y_{t}^{Q^{1}}  & = \sum_{t'=t+1}^{t+1+N} R_{t'} + \gamma \max_g Q(S_{t+1+N}, g | \mathbf{\theta_t}^{1})\\
        Y_t^{Q^2}  & = R_{t+1} + \gamma \max_a Q(S_{t+1}, a | \mathbf{\theta_t^2}, g)\\
    \end{split}
\end{equation}

\subsection{Reinforcement learning for hierarchical planning structure}

\begin{figure}[!t]
  \centering
  \includegraphics[width=0.9\columnwidth]{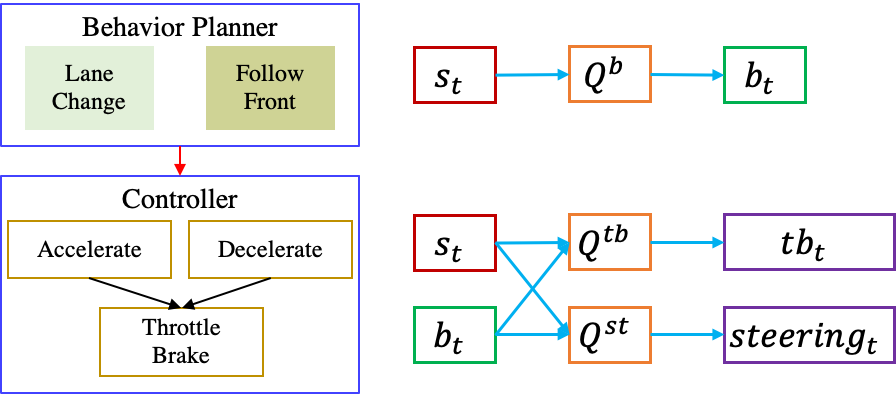}
  \caption{Hierarchical structure of planning system for two-layer setup. The networks are all fully connected.}
  \label{2layer_network}
\end{figure}

\begin{figure}[!t]
  \centering
  \includegraphics[width=0.9\columnwidth]{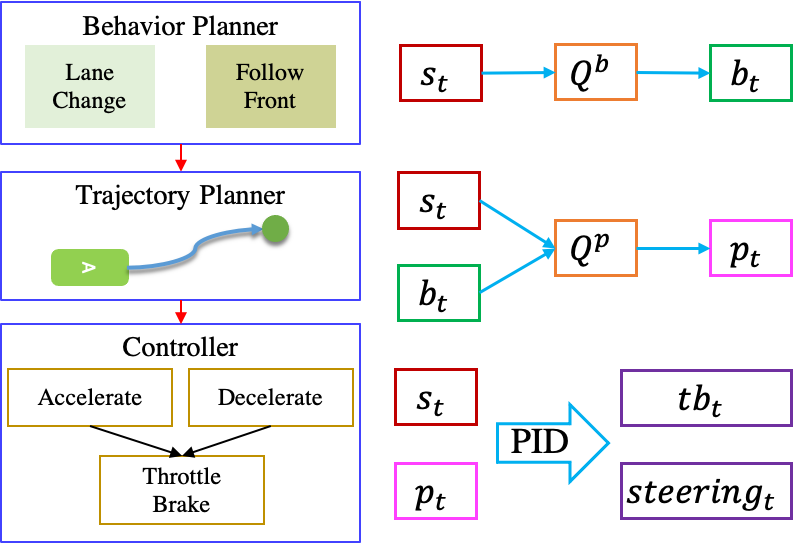}
  \caption{Hierarchical structure of planning system for three-layer setup. The networks are all fully connected.}
  \label{3layer_network}
  \vspace{-0.5cm}
\end{figure}

We propose HybridHRL, which can be applied to both two-layer and three-layer planning structures. For the two-layer structure, we use three fully connected networks, one to get an optimal behavior decision, and the other two to output the controller actions, which are throttle, brake, and steering angle. Figure \ref{2layer_network} shows the network structure for the two-layer setup. Similarly, for the three-layer setup, there are only two networks, one to get an optimal behavior decision, and the other to get the corresponding trajectory points that the ego car intends to trace. After that, a PID controller is used to optimize the lower-level controller system. Figure \ref{3layer_network} shows the hierarchical structure for the three-layer setup.

\subsection{Adjusted heuristic exploration}

During the reinforcement learning training process, some previous work \cite{hester2017deep} \cite{hester2017learning} \cite{nair2018overcoming} uses expert data to help the agent learn the optimal policy by embedding the difference between expert data and learnt behavior into the loss function. In our work, based on the original $\epsilon$-greedy method, we directly applied the heuristic rules enumeration policy to generate actions during the exploration period for training. Based on the exploration probability value $\epsilon$, for some epochs the action will explore according to the heuristic-based rules enumeration policy. Meanwhile, we adjust the decay rate for $\epsilon$ according to changing total reward of the current epoch. When the average total reward is higher than a period of previous epochs, $\epsilon$ is decreased during the training process. Otherwise, $\epsilon$ will increase to favor exploration over exploitation.

\begin{algorithm}[!t]\footnotesize
    \caption{HybridHRL for three-layer planning system}
    \label{algo_hybridhrl}
    \begin{algorithmic}[1]
    \Procedure{HybridHRL()}{}
        \State Initialize behavior-layer and trajectory-layer network $Q^b$,  $Q^p$ with weights $\theta^{b}$, $\theta^{p}$ and the target behavior and trajectory network ${Q}^{b'}$, ${Q}^{p'}$ with weights $\theta^{b'}$,  $\theta^{p'}$. 
        \State Construct an empty replay buffer $\mathbf{B}$ with max memory length $l_B$.
        \State $\epsilon = 1$, $k$ is a predefined training period number.
        \For {$e \gets 0$ to $E$ training epochs}
            \State Get initial states $s_0$.
            \While {$s$ is not the terminal state}
                \State Select behavior decision $B_t$ and $P_t$ based on $AdjustedHeuristicExploration()$.
                \State Apply PID controller to trace the trajectory point $P_t$ in simulation to get corresponding throttle, brake and steering angle and results in the next state $S_{t+1}$.
                \State $r^b_{t+1}, r^p_{t+1} = HybridReward(S_t, B_t, P_t)$.
                \State Store transition $T$ into $\mathbf{B}$: $T = \left\{S_t, B_t, P_t, r^b_{t+1}, r^p_{t+1}, S_{t+1} \right\}$.
            \EndWhile
            \State $R_e = \sum_t r_t$ \\
            \If {$\sum_{e-4k}^{e-2k} R_e < \sum_{e-2k}^{e} R_e$ }
                \State $\epsilon = \eta \epsilon$, $\eta \in [0, 1]$
            \Else  
                \State $\epsilon = \epsilon / \eta $, $\eta \in [0, 1]$
            \EndIf
            \State Train the buffer.
        \EndFor
    \EndProcedure
    \end{algorithmic}
\end{algorithm}

\begin{algorithm}[!t]\footnotesize
    \caption{Adjusted Heuristic Exploration}
    \label{algo_heuristic_explore}
    \begin{algorithmic}[1]
    \Procedure{AdjustedHeuristicExploration()}{}
        \If {$(e / k) \text{ mod } 2 = 0$ }
            \If {$random() > \epsilon$}
                \State $action = FollowHeuristicRule()$
            \Else
                \State $action = FollowHeuristicRule() + \mathcal{N}(\mu, \sigma)$
            \EndIf
        \Else
            \State $action = \arg\max_{action} Q(state, action)$
        \EndIf
        \Return action
    \EndProcedure
    \end{algorithmic}
\end{algorithm}

\begin{algorithm}[!t]\footnotesize
    \caption{Hybrid Reward Mechanism}
    \label{algo_hr}
    \begin{algorithmic}[1]
    \Procedure{HybridReward()}{}
        \State Penalize $r_t^o$ and $r_t^a$ for regular step penalties (e.x.: time penalty).
        \For {$\delta$ in sub-goals candidates}
            \If {$\delta$ fails}
                \If {option $o_t == \delta$}
                    \State Penalize option reward $r_t^o$
                \Else
                    \State Penalize action reward $r_t^a$
                \EndIf
            \EndIf
        \EndFor
        \If {task success (all $\delta$ success)}
            \State Reward both $r_t^o$ and $r_t^a$.
        \EndIf
    \EndProcedure
    \end{algorithmic}
\end{algorithm}

Algorithm \ref{algo_hybridhrl} describes the main training approach for a three-layer autonomous vehicle planning system. Algorithm \ref{algo_heuristic_explore} describes the method flow of adjusted heuristic exploration and Algorithm \ref{algo_hr} is the hybrid reward mechanism that is applied to calculate the hierarchical rewards for the training process.

\section{Experiments}

In this section, we show the results of applying HybridHRL to both two-level and three-level planning systems in some urban scenarios. Meanwhile, we compare our algorithms with heuristic-based rules enumeration policies, as well as previous RL approaches. We tested our algorithm in MSC’s VIRES VTD (Virtual Test Drive) simulator.

\subsection{Scenarios}

\begin{figure}[!t]
  \centering
  \includegraphics[width=\columnwidth]{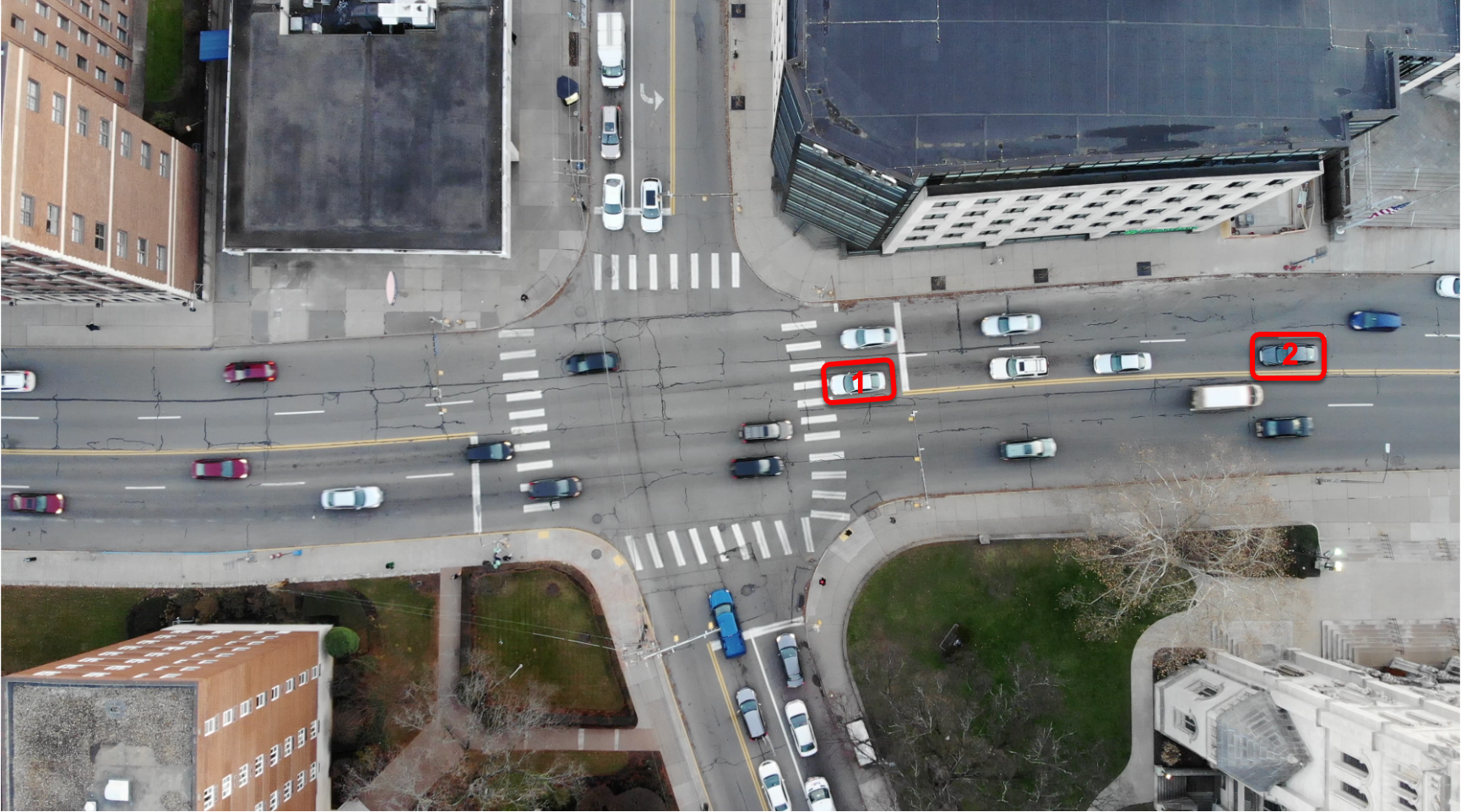}
  \caption{Scenarios of human drivers who have strong intentions to get through an urban intersection}
  \label{fig_scenario}
\end{figure}

Based on traffic data collected at an urban intersection in Pittsburgh, PA using UrbanFlow \cite{qiao2020human}, we notified two important cases which need strong human intention to get through. Figure \ref{fig_scenario} shows a snapshot of the two cases we will consider in the experiment part:

\begin{enumerate}
    \item For Car 1: The white car is trying to turn left at a green light. However, in Pittsburgh, most green traffic lights control all the vehicles either intending to go straight or turn left. As a result, the white car is blocked by vehicles going straight approaching from the opposite direction. When training the policy in simulation, the initial positions of the ego agent and other vehicles in the scene are all randomly assigned. There is no vehicle in front of the ego agent initially.
    
    \item For Car 2: The black car is trying to go straight at a green light. But because the white car described above is blocked, the black car is blocked as well. As a result, the driver needs to make a lane change in order to traverse the intersection during the current green light. When training the policy in the simulation, the initial positions of the ego agent and other vehicles in the scene are all randomly assigned. Either the left lane or right lane is blocked randomly. Sometimes, there is no lane blocked and the ego vehicle is not required to perform a lane-change behavior.

\end{enumerate}
In the attached video, we explain the two scenarios in detail.

\begin{table*}[!t]
\caption{Results comparisons between heuristic-based rules-enumeration and HybridHRL for left-turn scenario while applying two-layer planning system}
\label{table_ltresult}
\begin{center}
\begin{tabular}{|c||c c c|c|c c c c|}
\hline
& \multicolumn{3}{c|}{Rewards} & \multirow{2}{*}{Step}  & \multicolumn{4}{ c|}{ Performance Rate}\\
\cline{2-4}\cline{6-9}
& Behavior Reward &  Throttle Reward  & Steering Reward  & & Collision & Out of Road & Timeout & Success\\
\hline
Rule & 3.64 & -16.50 & 0.48 & 249 & 19.35\% & 11.83\% & 17.20\% & 51.61\%\\
HybridHRL &  82.98 & 71.97 & 87.37 & 52 & 0.8\% & 0.4\% & 0.0\% & 98.8\% \\
\hline
\end{tabular}
\end{center}
\end{table*}

\subsection{MDP Transitions}

\begin{figure}[!t]
  \centering
  \includegraphics[width=0.8\columnwidth]{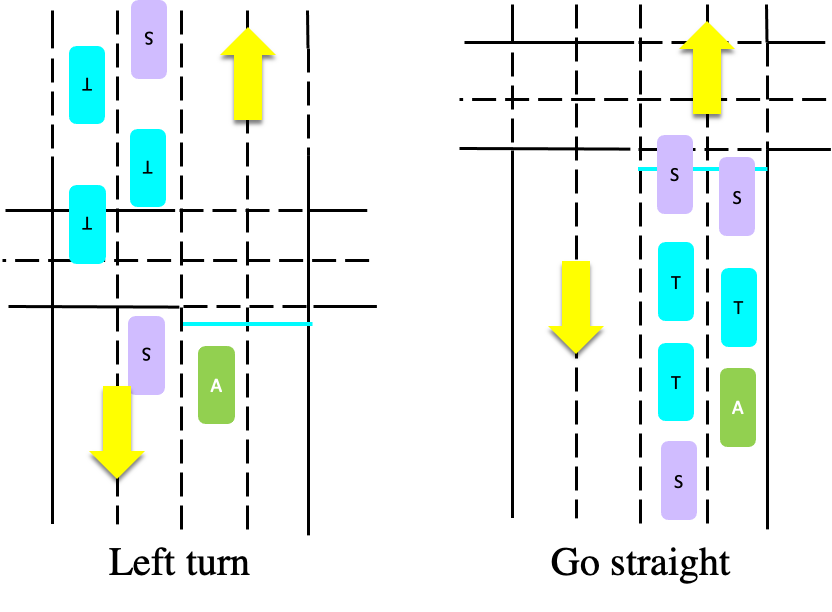}
  \caption{Green objects with letter $A$ are the ego agent for two scenarios. Cyan objects with letter $T$ are the selected target vehicles whose features are included in the state space.}
  \label{fig_tv}
  \vspace{-0.5cm}
\end{figure}

\subsubsection{State}

For each of the two scenarios, we choose three nearest target vehicles to be included in the state space. Figure \ref{fig_tv} shows the ego agent and target vehicles in different scenarios. The feature included in the state space can be categorized into two parts:
\begin{itemize}
    \item Ego agent features:
        1) velocity $v$, acceleration $a$, heading angle $h$; 2) lateral $d_{interx}$ and longitudinal $d_{intery}$ distance related to the center of the intersection; 3) time spent waiting at the intersection $t$;
     \item Target vehicle features:
        1) velocity $v$; acceleration $a$; heading angle $h$; 2) relative distance $d_a$ to the ego agent; 3) time to collision $ttc$ corresponding to the ego agent.
\end{itemize}

\subsubsection{Action}
\begin{itemize}
    \item High-level behavior planner:
        \begin{itemize}
            \item Left-turn scenario: \textit{LaneChange} or \textit{FollowFrontVehicle}.
            \item Go-straight scenario: \textit{LeftTurn} or \textit{Wait}.
        \end{itemize}
    \item Intermediate-level trajectory planner: the waypoint the agent intends to follow.
    \item Low-level controller: throttle or brake, steering angle
\end{itemize}

\subsubsection{Reward function}
The reward function can be categorized into two parts:
\begin{itemize}
    \item For each step:
        1) Time penalty; 2) Unsmoothness penalty if jerk is too large; 3) Unsafe penalty; 4) Penalty for deviating from the lane center when doing lane keeping. 
    \item For the termination conditions
        1) Collision penalty; 2) Timeout penalty. For example, the agent is stuck or waits too long to move forward; 3) Out of road penalty; 4) Success reward. 
\end{itemize}

\subsection{Results}

\begin{table*}[!t]
\caption{Results comparisons between heuristic-based rules-enumeration and HybridHRL for go-straight scenario}
\label{table_lcresult}
\begin{center}
\begin{tabular}{|c||c c c|c|c c c c|}
\hline
& \multicolumn{3}{c|}{Rewards} & \multirow{2}{*}{Step}  & \multicolumn{4}{ c|}{ Performance Metrics}\\
\cline{2-4}\cline{6-9}
& Behavior Reward  &  Throttle Reward & Steering Reward  & & Collision & Out of Road & Timeout & Success\\
\hline
Rule & 34.33 & 32.43 & 40.84 & 108 & 19.60\% & 0.0\% & 1.2\% & 79.8\%\\
HybridHRL 2-layer &  52.37 & 47.12 & 50.26 & 157 & 11.6\% & 3.7\% & 0.0\% & 84.7\% \\
HybridHRL 3-layer &  82.31 & 76.32 & 77.53 & 157 & 4.2\% & 0.2\% & 0.0\% & 95.6\% \\
\hline
\end{tabular}
\end{center}
\vspace{-0.5cm}
\end{table*}

% In this section, we describe the results for the proposed methods.

\subsubsection{Left-turn scenario}

\begin{figure}[!t]
  \centering
  \includegraphics[width=0.9\columnwidth]{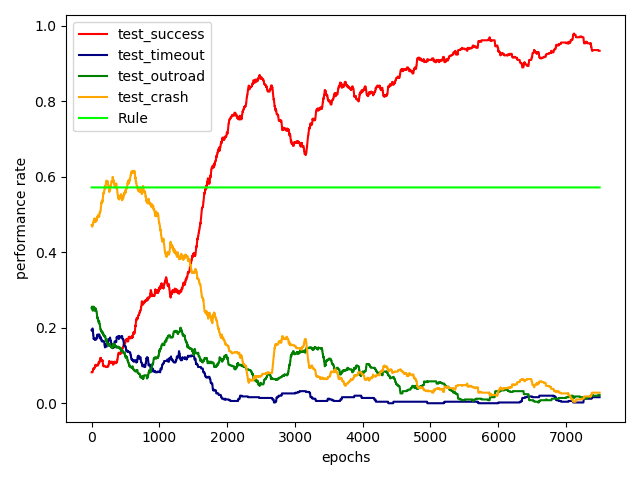}
  \caption{Performance rate for left-turn scenario during training process. The rate is an average performance for 500 test cases without explorations of the action space}
  \label{fig_lt_succ}
  \vspace{-0.5cm}
\end{figure}

\begin{figure}[!t]
  \centering
  \includegraphics[width=0.9\columnwidth]{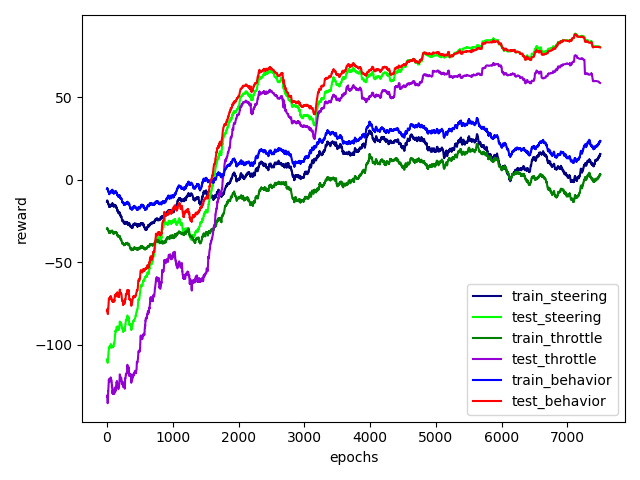}
  \caption{Reward for left-turn scenario during training process. The reward is an average performance for 500 cases. The train$\_$X results mean the actions are selected with adjusted heuristic exploration. The test$\_$X results mean the actions are selected directly through the network.}
  \label{fig_lt_reward}
  \vspace{-0.5cm}
\end{figure}

In Table \ref{table_ltresult}, we compare HybridHRL to the heuristic-based rules-enumeration policy which is based on predicting the difference of the time when arriving the potential collision point between the target vehicles and the ego agent. In addition to the great improvement of the performance rate, for reward evaluated from each level, we found out that both behavior and controller work better than rule-based method. In Figure \ref{fig_lt_succ} and Figure \ref{fig_lt_reward}, we can see the performance metrics and reward changing during the training process.

\begin{figure}[!t]
  \centering
  \includegraphics[width=\columnwidth]{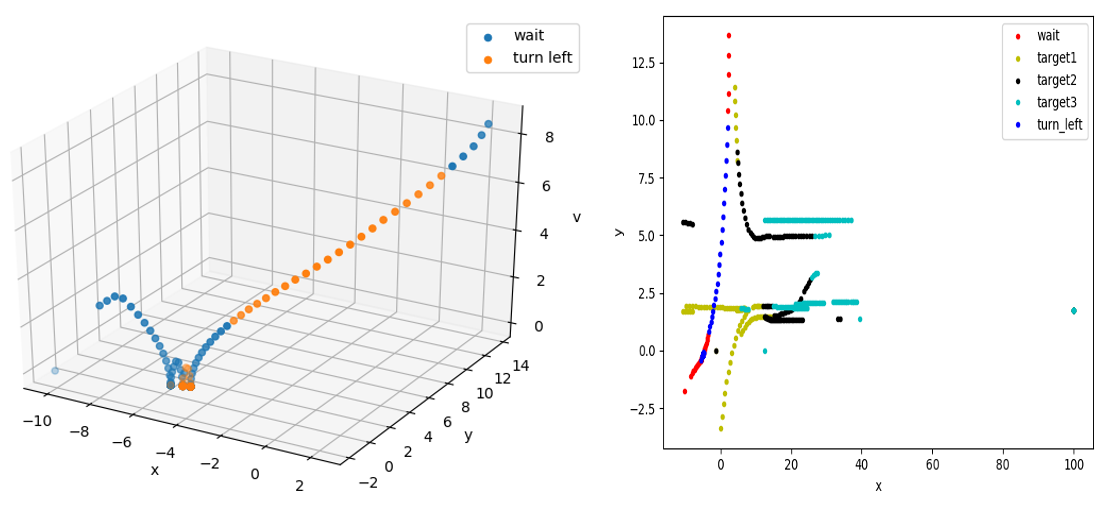}
  \caption{Heuristic-based rules-enumeration policy for left turn while encountering approaching vehicles from the opposite direction.}
  \label{fig_lt_rule_v}
  \vspace{-0.5cm}
\end{figure}

\begin{figure}[!t]
  \centering
  \includegraphics[width=\columnwidth]{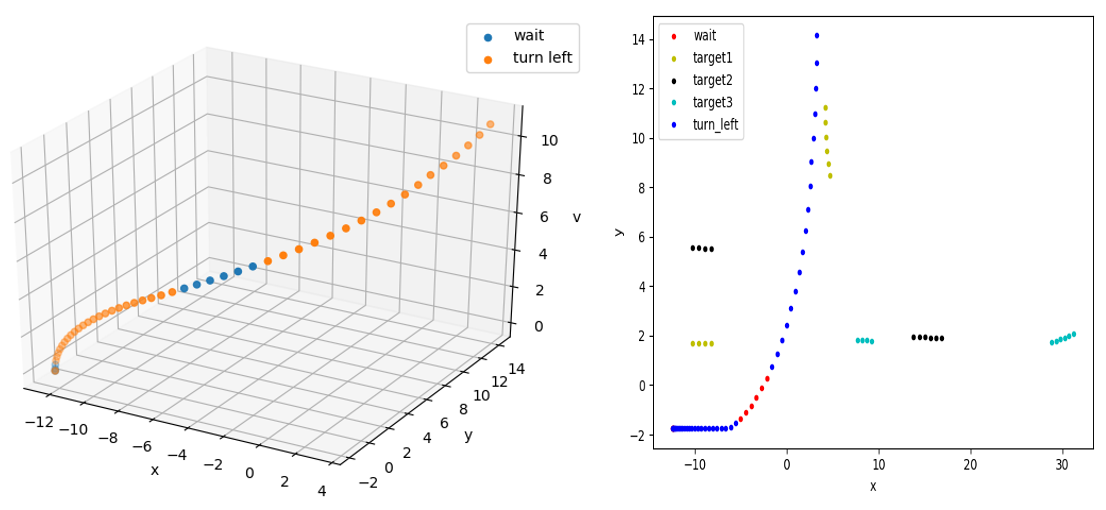}
  \caption{HybridHRL for left turn while encountering approaching vehicles from the opposite direction.}
  \label{fig_lt_hrl_v}
  \vspace{-0.5cm}
\end{figure}

For the same environment setup, we compare the behavior planner and velocity profiles for the rule-based method and HybridHRL. We set a constant time-to-collision threshold based on the velocity and the relative distance of the target vehicle for the rule-based methods and tune the threshold in order to get a relative high average reward for a set of test cases. But due to the high density of the traffic in the scene, the time-to-collision threshold is not robust to various situations. An obvious drawback is that the behavior decision of the rule-based method waits a much longer time than HybridHRL, which can quickly traverse the intersection with fewer steps. Figure \ref{fig_lt_rule_v} and \ref{fig_lt_hrl_v} compare the behavior-level decisions as well as the velocity profiles for the rule-based method and HybridHRL. In the 3D plots, the rule-based method is more likely to choose \textit{WAIT} during the left turn, which makes the ego agent slow down frequently. In the figures on the right, rule-based ego agent faces more target approaching vehicles than HybridHRL due to its waiting behavior.

\subsubsection{Go-straight scenario}

\begin{figure}[!t]
  \centering
  \includegraphics[width=\columnwidth]{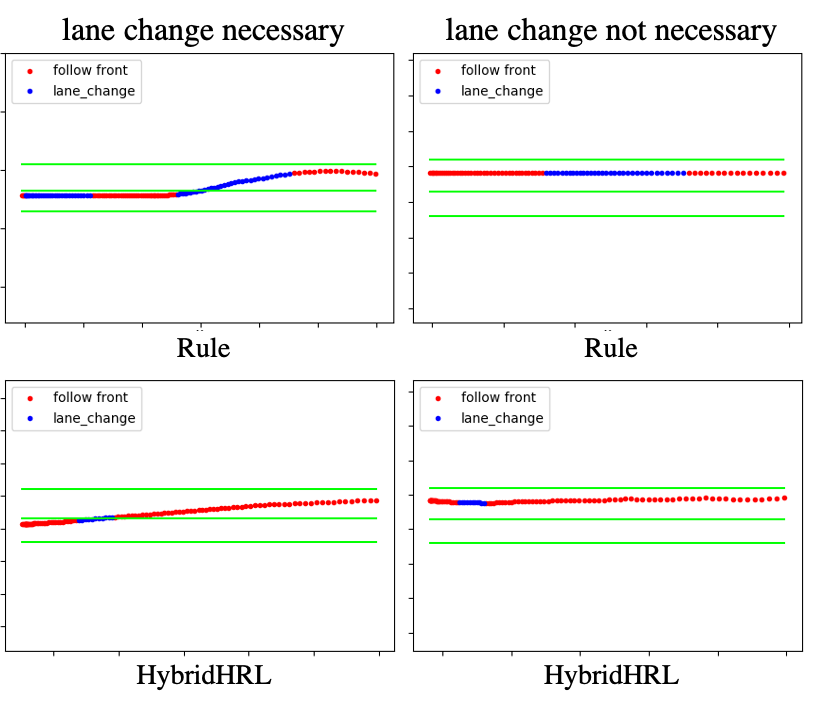}
  \caption{Behavior planner visualization for go-straight cases. For some situations when the no lane is blocked, the lane-change behavior is unnecessary. Different color dots show the corresponding behavior decisions that are selected.}
  \label{fig_lc_example}
  \vspace{-0.5cm}
\end{figure}

For the two-layer planning system, we visualize the behavior planner in Figure \ref{fig_lc_example}. The steering controller did not work well, especially for some cases that did not need lane change. When the ego agent moved forward, it could not stay centered in the lane. As a result, we introduced the three-layer planning system so that the trajectory planner can help to stabilize the controller.

In Table \ref{table_lcresult}, we compare the HybridHRL and heuristic-based rules-enumeration policy. For HybridHRL, we tested on both the two-layer and three-layer planning system. When evaluating the reward for the three-layer planning system, it shares the same reward mechanism with the two-layer planning system. 

\begin{figure}[!t]
  \centering
  \includegraphics[width=0.9\columnwidth]{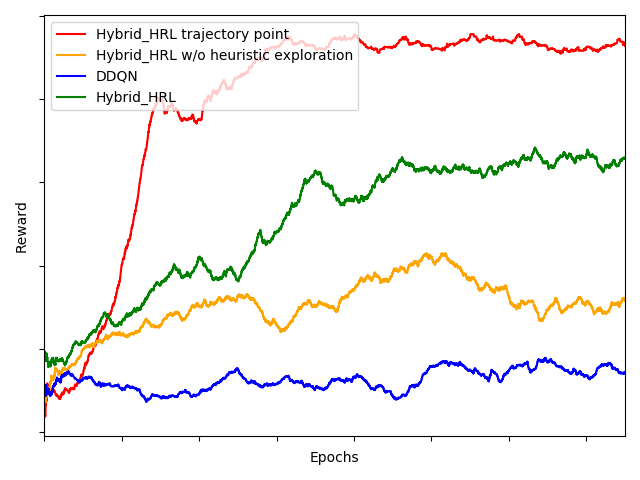}
  \caption{Training results of different RL algorithms for go-straight scenario.}
  \label{fig_comparison}
  \vspace{-0.5cm}
\end{figure}

Moreover, we tested the scenario for other reinforcement learning algorithms and visualized the training process in Figure \ref{fig_comparison}. For the Double DQN (DDQN) method, we used $\epsilon$-greedy exploration and no hybrid reward mechanism was applied. For HybridHRL w/o heuristic exploration, we applied the hybrid reward mechanism technique and the HybridHRL is for a two-layer planning system. The HybridHRL trajectory point is for the three-layer planning system, which also performs the best compared to the other methods.

For both scenarios we include the dynamic results in the video\footnote{\url{https://youtu.be/Wn3o0PwuVes}}. The video illustrates the planner results for both HybridHRL and rule-based decisions.

\section{CONCLUSIONS}

In this work, we successfully applied deep reinforcement learning to a broader range of challenging urban intersection scenarios. For the proposed method, we create a hierarchical RL structure with hybrid reward mechanism in order to deal with a multi-layer planning system for autonomous vehicles. Moreover, we applied a heuristic-based rules-enumeration policy during the exploration process in training to improve the training speed. Our algorithm shows superior performance to that of previous methods using various metrics.

\section*{Acknowledgments}

The authors would like to thank MSC Software for allowing us the use of VIRES Virtual Test Drive (VTD). The authors also would like to thank Weikun Zhen for expertly piloting the drone in the urban environment to collect the UrabnFlow data.

% \addtolength{\textheight}{-11cm}   % This command serves to balance the column lengths

%%%%%%%%%%%%%%%%%%%%%%%%%%%%%%%%%%%%%%%%%%%%%%%%%%%%%%%%%%%%%%%%%%%%%%%%%%%%%%%%

%%%%%%%%%%%%%%%%%%%%%%%%%%%%%%%%%%%%%%%%%%%%%%%%%%%%%%%%%%%%%%%%%%%%%%%%%%%%%%%%

%%%%%%%%%%%%%%%%%%%%%%%%%%%%%%%%%%%%%%%%%%%%%%%%%%%%%%%%%%%%%%%%%%%%%%%%%%%%%%%%

%%%%%%%%%%%%%%%%%%%%%%%%%%%%%%%%%%%%%%%%%%%%%%%%%%%%%%%%%%%%%%%%%%%%%%%%%%%%%%%%

\bibliography{behavior}

\begin{thebibliography}{10}
\providecommand{\url}[1]{#1}
\csname url@samestyle\endcsname
\providecommand{\newblock}{\relax}
\providecommand{\bibinfo}[2]{#2}
\providecommand{\BIBentrySTDinterwordspacing}{\spaceskip=0pt\relax}
\providecommand{\BIBentryALTinterwordstretchfactor}{4}
\providecommand{\BIBentryALTinterwordspacing}{\spaceskip=\fontdimen2\font plus
\BIBentryALTinterwordstretchfactor\fontdimen3\font minus
  \fontdimen4\font\relax}
\providecommand{\BIBforeignlanguage}[2]{{%
\expandafter\ifx\csname l@#1\endcsname\relax
\typeout{** WARNING: IEEEtran.bst: No hyphenation pattern has been}%
\typeout{** loaded for the language `#1'. Using the pattern for}%
\typeout{** the default language instead.}%
\else
\language=\csname l@#1\endcsname
\fi
#2}}
\providecommand{\BIBdecl}{\relax}
\BIBdecl

\bibitem{ttc}
D.~N. Lee, ``A theory of visual control of braking based on information about
  time-to-collision,'' \emph{Perception}, vol.~5, no.~4, pp. 437--459, 1976.

\bibitem{slot}
C.~R. Baker and J.~M. Dolan, ``Traffic interaction in the urban challenge:
  Putting boss on its best behavior,'' in \emph{2008 IEEE/RSJ International
  Conference on Intelligent Robots and Systems}.\hskip 1em plus 0.5em minus
  0.4em\relax IEEE, 2008, pp. 1752--1758.

\bibitem{ross2011reduction}
S.~Ross, G.~Gordon, and D.~Bagnell, ``A reduction of imitation learning and
  structured prediction to no-regret online learning,'' in \emph{Proceedings of
  the fourteenth international conference on artificial intelligence and
  statistics}, 2011, pp. 627--635.

\bibitem{zhang2016query}
J.~Zhang and K.~Cho, ``Query-efficient imitation learning for end-to-end
  autonomous driving,'' \emph{arXiv preprint arXiv:1605.06450}, 2016.

\bibitem{sun2018fast}
L.~Sun, C.~Peng, W.~Zhan, and M.~Tomizuka, ``A fast integrated planning and
  control framework for autonomous driving via imitation learning,'' in
  \emph{Dynamic Systems and Control Conference}, vol. 51913.\hskip 1em plus
  0.5em minus 0.4em\relax American Society of Mechanical Engineers, 2018, p.
  V003T37A012.

\bibitem{irl}
A.~Y. Ng, S.~J. Russell \emph{et~al.}, ``Algorithms for inverse reinforcement
  learning.'' in \emph{Icml}, vol.~1, 2000, p.~2.

\bibitem{chen2015deepdriving}
C.~Chen, A.~Seff, A.~Kornhauser, and J.~Xiao, ``Deepdriving: Learning
  affordance for direct perception in autonomous driving,'' in
  \emph{Proceedings of the IEEE International Conference on Computer Vision},
  2015, pp. 2722--2730.

\bibitem{bojarski2016end}
M.~Bojarski, D.~Del~Testa, D.~Dworakowski, B.~Firner, B.~Flepp, P.~Goyal, L.~D.
  Jackel, M.~Monfort, U.~Muller, J.~Zhang \emph{et~al.}, ``End to end learning
  for self-driving cars,'' \emph{arXiv preprint arXiv:1604.07316}, 2016.

\bibitem{dorsaplanning}
D.~Sadigh, S.~Sastry, S.~A. Seshia, and A.~D. Dragan, ``Planning for autonomous
  cars that leverage effects on human actions.'' in \emph{Robotics: Science and
  Systems}, vol.~2.\hskip 1em plus 0.5em minus 0.4em\relax Ann Arbor, MI, USA,
  2016.

\bibitem{rl}
R.~S. Sutton and A.~G. Barto, \emph{Reinforcement learning: An
  introduction}.\hskip 1em plus 0.5em minus 0.4em\relax MIT press Cambridge,
  1998, vol.~1, no.~1.

\bibitem{dqn}
V.~Mnih, K.~Kavukcuoglu, D.~Silver, A.~Graves, I.~Antonoglou, D.~Wierstra, and
  M.~Riedmiller, ``Playing atari with deep reinforcement learning,''
  \emph{arXiv preprint arXiv:1312.5602}, 2013.

\bibitem{ddpg}
T.~P. Lillicrap, J.~J. Hunt, A.~Pritzel, N.~Heess, T.~Erez, Y.~Tassa,
  D.~Silver, and D.~Wierstra, ``Continuous control with deep reinforcement
  learning,'' \emph{arXiv preprint arXiv:1509.02971}, 2015.

\bibitem{doubledqn}
H.~Van~Hasselt, A.~Guez, and D.~Silver, ``Deep reinforcement learning with
  double q-learning,'' in \emph{Thirtieth AAAI conference on artificial
  intelligence}, 2016.

\bibitem{pomdpdecision}
S.~Brechtel, T.~Gindele, and R.~Dillmann, ``Probabilistic decision-making under
  uncertainty for autonomous driving using continuous pomdps,'' in \emph{17th
  International IEEE Conference on Intelligent Transportation Systems
  (ITSC)}.\hskip 1em plus 0.5em minus 0.4em\relax IEEE, 2014, pp. 392--399.

\bibitem{navigateintersection}
D.~Isele, A.~Cosgun, K.~Subramanian, and K.~Fujimura, ``Navigating
  intersections with autonomous vehicles using deep reinforcement learning,''
  \emph{arXiv preprint arXiv:1705.01196}, 2017.

\bibitem{occludednavigating}
D.~Isele, R.~Rahimi, A.~Cosgun, K.~Subramanian, and K.~Fujimura, ``Navigating
  occluded intersections with autonomous vehicles using deep reinforcement
  learning,'' in \emph{2018 IEEE International Conference on Robotics and
  Automation (ICRA)}.\hskip 1em plus 0.5em minus 0.4em\relax IEEE, 2018, pp.
  2034--2039.

\bibitem{maxq}
T.~G. Dietterich, ``The maxq method for hierarchical reinforcement learning.''
  in \emph{ICML}, vol.~98.\hskip 1em plus 0.5em minus 0.4em\relax Citeseer,
  1998, pp. 118--126.

\bibitem{rmax}
R.~I. Brafman and M.~Tennenholtz, ``R-max-a general polynomial time algorithm
  for near-optimal reinforcement learning,'' \emph{Journal of Machine Learning
  Research}, vol.~3, no. Oct, pp. 213--231, 2002.

\bibitem{hrl}
N.~K. Jong and P.~Stone, ``Hierarchical model-based reinforcement learning:
  R-max+ maxq,'' in \emph{Proceedings of the 25th international conference on
  Machine learning}.\hskip 1em plus 0.5em minus 0.4em\relax ACM, 2008, pp.
  432--439.

\bibitem{hdrl}
T.~D. Kulkarni, K.~Narasimhan, A.~Saeedi, and J.~Tenenbaum, ``Hierarchical deep
  reinforcement learning: Integrating temporal abstraction and intrinsic
  motivation,'' in \emph{Advances in neural information processing systems},
  2016, pp. 3675--3683.

\bibitem{parl}
W.~Masson, P.~Ranchod, and G.~Konidaris, ``Reinforcement learning with
  parameterized actions,'' in \emph{AAAI}, 2016, pp. 1934--1940.

\bibitem{hester2017deep}
T.~Hester, M.~Vecerik, O.~Pietquin, M.~Lanctot, T.~Schaul, B.~Piot, D.~Horgan,
  J.~Quan, A.~Sendonaris, G.~Dulac-Arnold \emph{et~al.}, ``Deep q-learning from
  demonstrations,'' \emph{arXiv preprint arXiv:1704.03732}, 2017.

\bibitem{nair2018overcoming}
A.~Nair, B.~McGrew, M.~Andrychowicz, W.~Zaremba, and P.~Abbeel, ``Overcoming
  exploration in reinforcement learning with demonstrations,'' in \emph{2018
  IEEE International Conference on Robotics and Automation (ICRA)}.\hskip 1em
  plus 0.5em minus 0.4em\relax IEEE, 2018, pp. 6292--6299.

\bibitem{hester2017learning}
T.~Hester, M.~Vecerik, O.~Pietquin, M.~Lanctot, T.~Schaul, B.~Piot,
  A.~Sendonaris, G.~Dulac-Arnold, I.~Osband, J.~Agapiou \emph{et~al.},
  ``Learning from demonstrations for real world reinforcement learning,'' 2017.

\bibitem{qiao2020human}
Z.~Qiao, J.~Zhao, J.~Zhu, Z.~Tyree, P.~Mudalige, J.~Schneider, and J.~M. Dolan,
  ``Human driver behavior prediction based on urbanflow,'' in \emph{2020 IEEE
  International Conference on Robotics and Automation (ICRA)}.\hskip 1em plus
  0.5em minus 0.4em\relax IEEE, 2020, pp. 10\,570--10\,576.

\end{thebibliography}
\bibliographystyle{IEEEtran}

\end{document}